\documentclass{article}
\usepackage{spconf,amsmath,graphicx}

\usepackage{comment}
\usepackage{booktabs}
\usepackage{placeins}
\usepackage{float}
\usepackage[usenames,dvipsnames]{xcolor}
\usepackage{subcaption}

\title{Analyzing the factors affecting usefulness of Self-Supervised Pre-trained Representations for Speech Recognition}
\name{Ashish Seth$^{1\star}$,
Vasista Sai Lodagala$^{1\star}$,
Sreyan Ghosh$^{1\star}$, S. Umesh$^{1}$ \thanks{\hspace*{-1mm}$^{\star}$These authors contributed equally to this work}}

\address{
  $^1$Speech Lab, Department of Electrical Engineering, IIT Madras, Chennai, India\\
  }
\begin{document}
%
\maketitle
\begin{abstract}
Representations from well-trained Self-supervised learning (SSL) models have become popular while building Automatic Speech Recognition (ASR) systems when the amount of labelled data is scarce.
Though widely used in practice, \emph{continued pre-training}, or pre-training an already pre-trained model on in-domain data, lacks systematic study. In this work, we first investigate how factors such as domain, language and size of the upstream SSL pre-training data affect the final downstream ASR performance. Finally, we investigate the continued pre-training paradigm and analyze how upstream model selection plays an essential role in this setup. Through thorough quantitative and qualitative analysis, we reveal that the performance of ASR systems is sensitive to the data used for SSL pre-training. Moreover, under extremely low-resource conditions ($\leq$ 10 hours of fine-tuning data), leveraging features from SSL models result in significant gains over traditional surface features like FBank (up to 29\% absolute WER improvement).

\end{abstract}
\vspace{-0.5em}
\section{Introduction}
\label{sec:intro}
\vspace{-0.5em}
One of the major problems with building efficient Automatic Speech Recognition (ASR) systems is that they are data-hungry \cite{chiu2018state}, and with the introduction of deep learning-based methods, this problem has amplified. Though English has more than 100k hours of human-transcribed data freely available online, for languages beyond English, data is scarce. Some languages even lack professional annotators and exist only in spoken form.
With researchers and businesses finding the true potential of ASR in various Natural Language Understanding (NLU) systems, there has been an increasing demand for building such systems in languages beyond English.

Though SSL models learn better task-agnostic features \cite{superb_bench}, they are not robust to changes in the domain between the pre-training and ASR fine-tuning data, as they get biased towards the domain of the unlabeled data \cite{hsu21_interspeech}.
This results in the need for huge amounts of unlabeled data with a high source similarity to the final downstream low-resource data in terms of accent, language, and domain. Also, as rightly pointed out by \cite{hannun2021history}, SSL would be more accessible if it could be trained with lesser compute.
\vspace{1mm}

{\noindent \textbf{Main Results.}} This paper builds on our curiosity to answer the following question. \emph{``Can we use existing models available online, pre-trained on thousands of hours of data, to improve performance on a rather low-resource ASR task with minimal pre-training and fine-tuning steps?''} and if so, \emph{``How would a shift in the domain, or language in our labeled fine-tuning data, compared to the original pre-training unlabeled data, affect the performance of our model for ASR?''}. Thus, building on the \emph{continued pre-training} paradigm, we use the 1000 hours of unlabeled data from the Interspeech GramVaani ASR challenge for pre-training purposes.  
We use SSL models pre-trained on varied domains and languages as upstream feature extractors.
Feature representations from these SSL models are then used for the downstream ASR task.
Our setup is very similar to \cite{chang2021exploration}, which was one of the first works to explore speech-based SSL pre-trained models as upstream feature extractors for raw speech. This pre-training and feature extraction paradigm has been relatively under-studied, and our work contributes considerably in exploring this better. Beyond achieving considerable improvements over our baselines, contradictory to \cite{hsu21_interspeech} where the analysis was done on a combined data setup, we do the first of its kind analysis on the \emph{continued pre-training} setup, which is more feasible in a real-world setting. Extensive experiments reveal that source pre-trained data matters for the target \emph{continued pre-training}, significantly affecting the final downstream ASR task performance.

\vspace{-0.5em}
\section{Related Work}
\label{sec:related_work}
\vspace{-0.5em}
A comprehensive study of SSL techniques on various Spoken Language Processing (SLP) downstream tasks can be found in \cite{superb_bench}. With research in this field gaining attention, it has been of utmost importance to analyze the properties of these learned features. We need to investigate the generalizability of these features towards domain, language, and expand their capabilities to build systems beyond English. Researchers have recently made reasonable efforts to analyze the intrinsic properties of features learned through SSL at various layers of the deep neural networks \cite{pasad2021layer,shah2021all}. \cite{hsu21_interspeech} demonstrates that wav2vec 2.0 tends to get biased towards the pre-training data, and that domain similarity with the unlabeled data helps the final downstream ASR task. Along similar lines, in this work, we try to analyze the effect of domain, multi-linguality, and dataset size of the unlabeled data on our low-resource downstream ASR task. Different from \cite{hsu21_interspeech}, we build on the \emph{continued pre-training} paradigm \cite{gururangan-etal-2020-dont} which we acknowledge is more practical in a real-world setting than pre-training from scratch by combining the data.

\vspace{-0.5em}
\section{Proposed Methodology}
\label{sec:methodology}

\subsection{Model Architecture}
\label{sec:arch}
\vspace{-0.5em}
Our experiments, including the baselines, use either Filter-Bank or wav2vec 2.0 models as the upstream feature extractors. We use the conformer-based \cite{gulati2020conformer} Encoder-Decoder model as our downstream, which takes features from our upstream as input and trains to learn the task of ASR.

\vspace{-0.75mm}

\subsection{Upstream}
\label{sec:upstream}
\vspace{-0.5em}
Previous to the self-supervised pre-training era, FBank-Pitch was among the most common low-level feature extractor choices for building efficient ASR systems \cite{liu2020mockingjay,liu2021tera}. Thus, as shown in Fig.\ref{fig:speech_production}, the 80-dimensional FBank-Pitch features extracted from raw audio are directly fed to the downstream model for training on the ASR task. With the recent dawn of SSL, the use of SSL pre-trained models trained using Masked Acoustic Modelling (MAM) has shown to be a better alternative to low-level features like FBank-Pitch \cite{chang2021exploration}. As seen in Fig.\ref{fig:speech_production}, our setup is similar to \cite{chang2021exploration}, where either FBank-pitch features or features from pre-trained SSL models are fed into the downstream model for the ASR task.
We resort to wav2vec 2.0 (LARGE) as the feature extractor for our model, and to ensure a fair comparison with the FBank-Pitch feature extractor, we do not fine-tune our upstream model parameters while training on ASR. We acknowledge that fine-tuning the upstream might benefit the ASR task and remains part of our future work. It is to be noted that Context representations from the transformer encoder are used as the upstream features from the wav2vec 2.0 models. The SSL pre-training details of wav2vec 2.0 are beyond this paper's scope, and we refer our readers to \cite{baevski2020wav2vec} for more details.
\vspace{-0.75mm}

\subsection{Downstream}
\label{sec:downstream}
\begin{figure}[t]
  \includegraphics[width=0.45\textwidth]{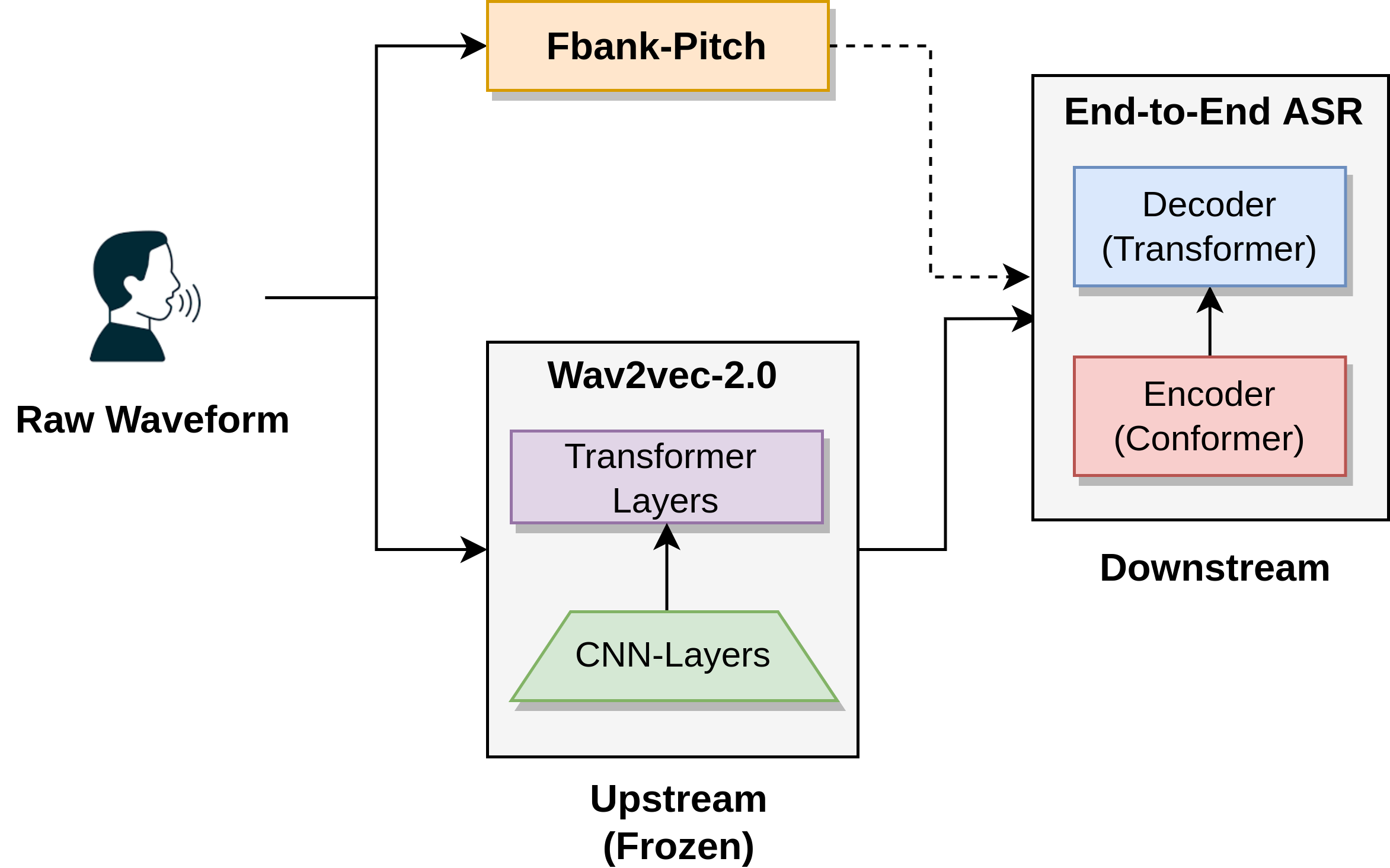}
  \caption{Illustration of the End-to-End ASR with various input speech representations.}
  \label{fig:speech_production}
\end{figure}

\vspace{-0.5em}
As shown in Fig.\ref{fig:speech_production}, our downstream ASR training is based on a joint CTC/attention-based encoder-decoder architecture \cite{watanabe2017hybrid} and we conduct all our downstream experiments using the ESPnet toolkit \cite{watanabe18_interspeech}. Our encoder is made up of 12 conformer encoder blocks. While transformer models are well known for capturing global information and data interactions, conformer blocks effectively capture local as well as global information \cite{gulati2020conformer}. Our downstream decoder uses 6 transformer blocks, and we train the models with a $0.4$ weight to CTC.

\vspace{-0.75em}
\section{Experiments}
\label{sec:exp}
\vspace{-0.5em}
\subsection{Datasets}
\label{sec:data}
\vspace{-0.5em}
We experiment with two ASR datasets covering 4 Indian Languages. We use the MSR (Microsoft Research) dataset, which was released as a part of the Low Resource Speech Recognition Challenge for Indian Languages \cite{srivastava18_sltu}. The dataset consists of 50 hours of speech data with transcriptions for Gujarati, Tamil, and Telugu amounting to 150 hours. For Hindi, we use the GramVaani dataset \cite{bhanushali22_interspeech}, which was released as a part of the Interspeech GramVaani ASR Challenge 2022. It is a corpus of 1108 hours of real-world, spontaneous telephone speech recordings in multiple dialects of the Hindi language. Of this 1108 hours of data, 1000 hours is unlabeled, 100 hours is the labeled training data, and 5 hours of development data. Blind test data of 3 hours has been released for evaluation. The audio files were recorded at different sampling rates ranging from 8 kHz to 48 kHz in the mp3 format.
For all the upstream pre-training / continued pre-training of the wav2vec-2.0 models, we use the 991 hours of the GramVaani unlabeled data for pre-training/continued pre-training and the rest 9 hours as a validation set.
The 100-hour labeled data is used for the downstream ASR training.

\vspace{-0.5em}
\subsection{Baselines}
\label{sec:baselines}
\vspace{-0.5em}
As mentioned in section \ref{sec:downstream}, all of the experiments have been carried out with the same downstream model configuration, the structural details of which have been mentioned in section \ref{sec:setup}. The baseline models differ from the rest of the systems in terms of the upstream used.
For the first baseline system, we use 80-dimensional FBank-Pitch features for training the downstream encoder-decoder model.
As the second baseline system, we pre-train a wav2vec-2.0 model from scratch on the GramVaani unlabeled data and use it as the upstream feature extractor to train the downstream encoder-decoder model.

\vspace{-0.5em}
\subsection{Experimental Setup}
\label{sec:setup}

{\noindent \textbf{Upstream Configuration.}} The wav2vec 2.0 models used as the upstream feature extractors, have been pre-trained on data from different languages, domains, and datasets of various sizes. We use wav2vec 2.0 LARGE models (24 transformer layers) for all our experiments. Following are the descriptions of the SSL models used, and the datasets they have been pre-trained on:

{\noindent \textbf{LL}} : The model is pre-trained on 60k hours of the Libri-Light dataset which is a corpus of English audiobook data \cite{baevski2020wav2vec}.

{\noindent \textbf{LL+CV+SF}} : The model is pre-trained on pooled multi-domain data sourced from Libri-Light, CommonVoice \cite{ardila2019common}, Switchboard \cite{godfrey1992switchboard} and Fisher datasets.

{\noindent \textbf{XLSR-128}} : The model is pre-trained on 436k hours of unlabeled speech data from 128 different languages \cite{babu2021xlsr}.

{\noindent \textbf{IndicW2V}} : The model is pre-trained on 17k hours of multilingual data from 40 Indian languages \cite{javed2021towards}. Also, the data used has been sourced from varied domains, including education, news, technology, and finance.


The 1024-dimensional representations extracted from these models, are passed through a linear layer to obtain 80-dimensional features, which are then fed to the downstream model for the End-to-End ASR training. 
\vspace{0.5mm}

{\noindent \textbf{Downstream Configuration.}} Our downstream model is a conformer-based encoder-decoder model. Our encoder has 12 conformer blocks with eight attention heads and outputs a $512$-dimensional representation for each frame which it passes to the decoder. Our decoder has 6 transformer blocks with eight attention heads. While decoding, we analyze the effect of various upstreams independent of the language model. During inference, we perform beam search decoding with a beam size of 20. All the optimal hyper-parameters have been found via grid-search.
\vspace{0.5mm}

{\noindent \textbf{Continued Pre-training (CP).}} To better adapt the above mentioned SSL pre-trained models to the downstream tasks, we allow these models to continue pre-training on the GramVaani unlabeled data.
First introduced in \cite{gururangan-etal-2020-dont}, \emph{continued pre-training} (CP) has proven to be quite effective for unsupervised domain adaptation in NLP. All the pre-training and fine-tuning experiments have been conducted using the Fairseq \cite{ott2019fairseq} and ESPnet toolkits \cite{watanabe18_interspeech} respectively, with 4 A-100 GPUs used for pre-training and 2 A-100 GPUs used for fine-tuning. All the hyper-parameters for pre-training have been borrowed from the original setting in \cite{baevski2020wav2vec, ott2019fairseq}.


\begin{figure}
\centering
\begin{subfigure}[b]{0.58\textwidth}
   \includegraphics[width=\linewidth]{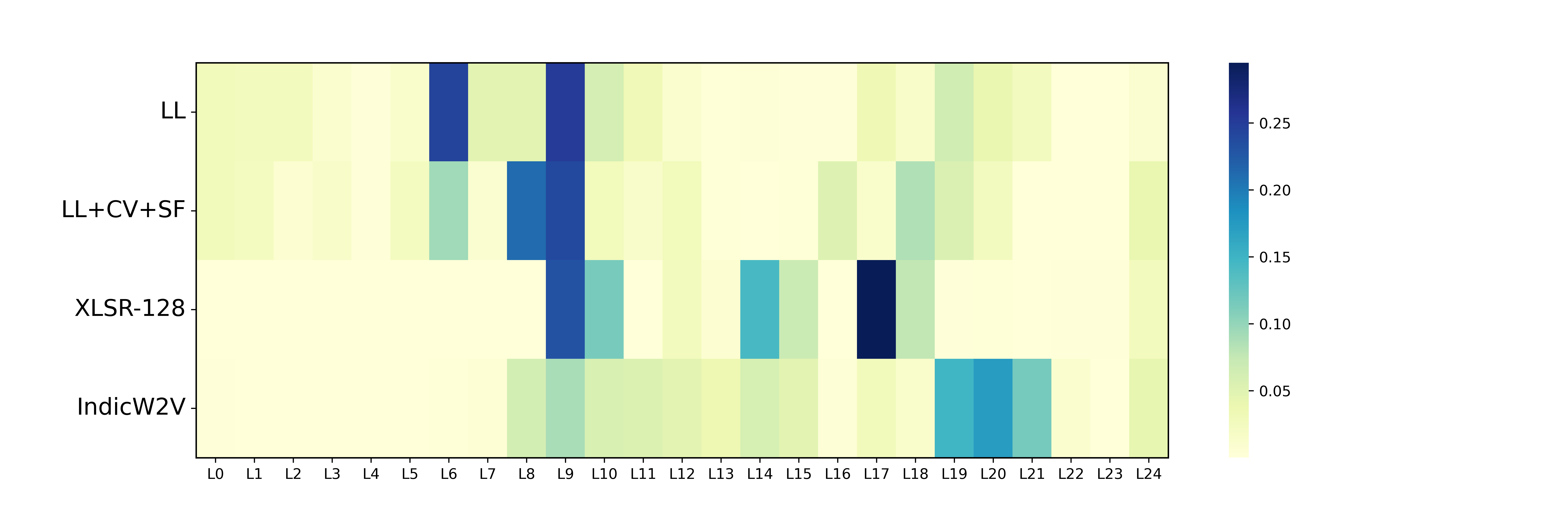}
   \caption{Feature extractor \emph{before} continued pre-training}
   \label{fig:Ng1} 
\end{subfigure}

\begin{subfigure}[b]{0.58\textwidth}
   \includegraphics[width=\linewidth]{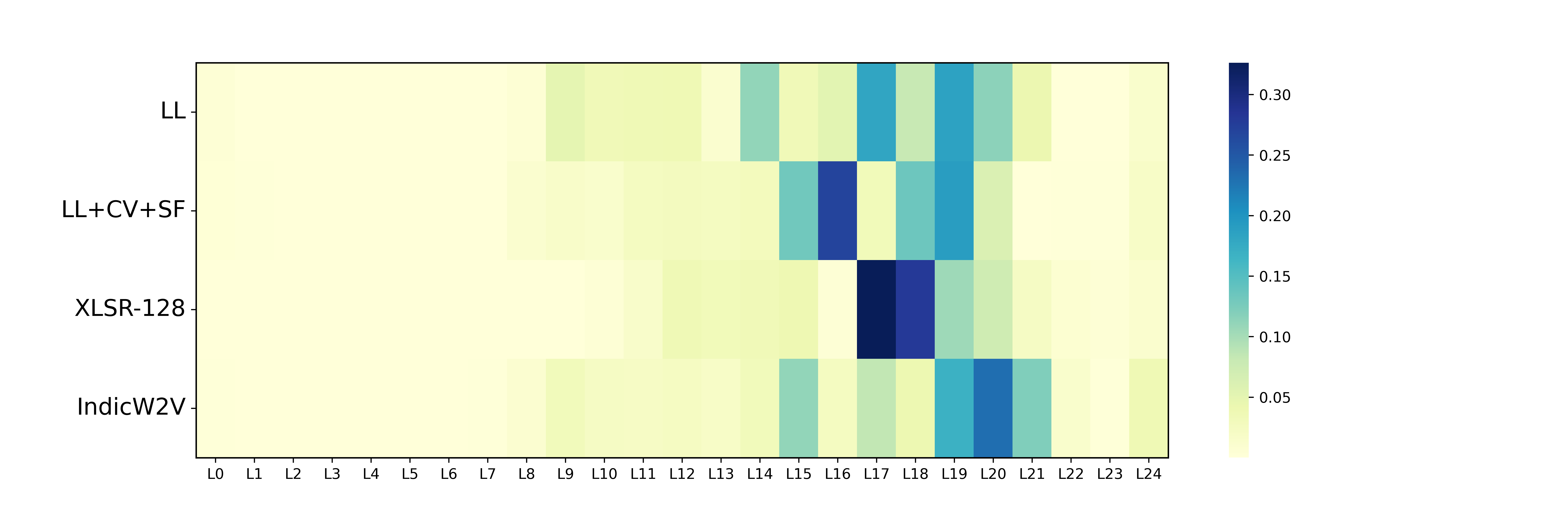}
   \caption{Feature extractor \emph{after} continued pre-training}
   \label{fig:Ng2}
\end{subfigure}

\caption[Two numerical solutions]{(a), (b) shows Layer-wise attention weights for downstream fine-tuning with different wav2vec 2.0 feature extractors. ``L0'' is the convolutional feature encoder.}
\label{fig:layer_heatmap}
\end{figure}

\begin{table*}[t]
\setlength{\tabcolsep}{8pt}
\centering
  \caption{Performance of various E2E-ASR models with different upstream feature extractors}

  \label{tab:example_1}

  \begin{tabular}{l l l l l l l l l l l l l}
    \toprule
    \textbf{Feature Extractor} & & \multicolumn{2}{c}{\textbf{Hindi}} & & \multicolumn{2}{c}{\textbf{Gujarati}} & & \multicolumn{2}{c}{\textbf{Tamil}} & & \multicolumn{2}{c}{\textbf{Telugu}}\\
    & & dev & test & & dev & test & & dev & test & & dev & test \\
    
    \textbf{Baseline} & & & & & & & & & & & & \\
    \toprule
    1. Fbank-Pitch & & 34.2 & 33.7 & & 27.8 & 35.3 & & 30.1 & 29.8 & & 32.8 & 32.9 \\
    2. W2V-GV & & 32.4 & 32.3 & & 25.6 & 34.4 & & 29.9 & 29.3 & & 32.0 & 31.9 \\
    \hline
    \textbf{Pre-Trained Models} & & & & & & & & & & & & \\
    \toprule
    3. LL & & 35.0 & 34.4 & & 22.1 & 29.8 & & 28.6 & 28.4 & & 29.1 & 30.1 \\
    4. LL+CV+SF & & 34.3 & 34.2 & & 22.0 & 29.5 & & 28.2 & 27.9 & & 28.5 & 28.9 \\
    5. XLSR-128 & & 32.7 & 32.5 & & \textbf{21.7} & \textbf{28.5} & & \textbf{28.1} & \textbf{27.7} & & \textbf{28.3} & \textbf{28.8} \\
    6. IndicW2V & & 33.6 & 33.1 & & 21.8 & 29.0 & & 28.3 & 28.2 & & 29.5 & 30.1 \\
    \hline
    \textbf{Continued Pre-Trained (CP) Models} & & & & & & & & & & & & \\
    \toprule
    7. LL & & 29.7 & 29.8 & & 22.2 & 29.5 & & 29.0 & 28.4 & & 29.1 & 29.6 \\
    8. LL+CV+SF & & 29.1 & 28.9 & & 22.8 & 30.6 & & 29.6 & 29.6 & & 30.8 & 31.0 \\
    9. XLSR-128 & & \textbf{27.3} & \textbf{27.1} & & 24.5 & 32.8 & & 30.6 & 30.7 & & 31.4 & 32.1 \\
    10. IndicW2V & & 31.4 & 31.5 & & 24.0 & 32.8 & & 31.0 & 30.6 & & 31.6 & 32.2 \\
    \bottomrule
    
  \end{tabular}

\end{table*}

\begin{table}[t]
\setlength{\tabcolsep}{7pt}
    \centering
  \caption{Performance of various E2E-ASR models with different upstream feature extractors using 10h labeled data}

  \label{tab:example}

  \begin{tabular}{l l l l}
    \toprule
    \textbf{Feature Extractor} & & \multicolumn{2}{c}{\textbf{Hindi}} \\
    & & dev & test \\
    
    \textbf{Baselines} & & & \\
    \toprule
    1. Fbank-Pitch & & 75.7 & 74.5 \\
    2. W2V-GV & & 50.4 & 50.3 \\
    \toprule
    3. XLSR-128 & & 56.8 & 55.9 \\
    4. XLSR-128 (Continued pre-trained) & & \textbf{46.2} & \textbf{46.0} \\

    \bottomrule
    
  \end{tabular}

\end{table}

\vspace{-0.5em}
\section{Results and Analysis}
\label{sec:results}
\vspace{-0.5em}
Table \ref{tab:example_1} reports the \%WER results for all our experiments across the 4 Indian languages. 
W2V-GV indicates our second baseline system, where we pre-train the wav2vec 2.0 model from scratch on the GramVaani unlabeled data and use it as the upstream feature extractor.
XLSR-128 as our upstream feature extractor proves to be our best setup on Hindi. It outperforms our FBank-Pitch baselines by 1.5\%, and 1.2\% WER on the dev and test sets. However, after \emph{continued pre-training} XLSR-128 as a feature extractor outperforms the FBank-Pitch baselines by 6.9\%, and 6.6\% WER. 

Table \ref{tab:example} reports the \%WER results with the amount of fine-tuning data limited to 10 hours, sampled from the 100h GramVaani labeled data.
XLSR-128 with \emph{continued pre-training} which was the best performing feature extractor from Table \ref{tab:example_1} outperforms our FBank-Pitch baseline by an absolute 29.5\% WER and our W2V-GV baseline by 4.2\% WER, thus proving the benefits of SSL and continued pre-training in extreme low-resource labeled data training regimes. We next analyze how language, domain and dataset size factors affect the usefulness of features obtained from pre-trained and continued pre-trained SSL models.

{\noindent \textbf{Feature Extractor's layer importance}}: Since we use the weighted sum of features from all the layers of our upstream model for our downstream ASR fine-tuning, we try to analyze the importance given to each layer by our ASR task, for the model fine-tuned on the Hindi dataset. Fig.\ref{fig:Ng1} and Fig.\ref{fig:Ng2} show a pictorial representation of attention weights for each layer, for each upstream model, before and after CP respectively. As we see, before CP our task pays scattered attention across all layers based on the information contained in these respective layers. For example, in lines with the observation made by \cite{pasad2021layer}, that representation from layers 16-21 of a wav2vec-2.0 LARGE model has the highest phonetic identity for IndicW2V which has seen the highest amount of Hindi data even prior to CP. ASR fine-tuning therefore draws the most information from these layers. Contrary to this, for LL and LL+CV+SF, which were pre-trained entirely on a different language, our task draws no information from these layers. However, for all the models after CP, our downstream ASR task gives more attention to these layers than before, which results in the model learning more language-specific phonetic properties.
\vspace{0.5mm}

{\noindent \textbf{Upstream data volume and domains}}: As shown in Table \ref{tab:example_1}, XLSR-128, which is pre-trained on more than 436K hours of data, outperforms all the other pre-trained models. After continued pre-training, a similar trend is observed for the Hindi labeled data. Compared to other pre-trained models, XLSR-128 is pre-trained on huge amounts of unlabeled data, giving a better initialization before continued pre-training. Also from Table \ref{tab:example_1}, LL+CV+SF  performs better than LL, and XLSR-128 performs better than IndicW2V across all the languages. From this we conclude that models pre-trained on more domains learn better domain-agnostic features.
\vspace{0.5mm}

{\noindent \textbf{Effectiveness of SSL features in low-resource data regimes}}: From the 100hours GramVaani labeled data, we randomly sample 10hours of data for fine-tuning. As shown in Table \ref{tab:example}, the continued pre-trained XLSR-128 model results in an average absolute improvement of 29\% WER from our FBank-Pitch baseline on the dev and the test set.
From Table \ref{tab:example_1} we observe that using SSL features benefitted relatively low resource languages like Gujarati, Tamil, and Telugu (50h labeled data) compared to Hindi (100h labeled data).
Therefore, using SSL representations for low-resource fine-tuning proves to be more effective than the FBank-Pitch features.
\vspace{0.5mm}

{\noindent \textbf{Continued pre-training on a different language doesn't help}}: For multilingual models such as XLSR-128, we observe an average hit of 3.5\%, 2.8\%, and 3.2\% of WER on Gujarati, Tamil, and Telugu data respectively, on the dev and test sets after continued pre-training on the unlabeled Hindi data. We observe a similar trend with IndicW2V. We hypothesize that for \emph{multilingual models, continued pre-training on a different language leads to catastrophic forgetting of the in-domain knowledge} \cite{kaushik2021understanding}. 
As mentioned in \cite{babu2021xlsr}, among the 128 languages, XLSR-128 is pre-trained on 37 hours of Gujarati, 118 hours of Tamil, and 62 hours of Telugu. When we perform continued pre-training on a different language, in the process of adapting to the new domain, the model fails to retain the knowledge from its multilingual pre-training, leading to catastrophic forgetting.

\vspace{-0.5em}
\section{Conclusion}
\vspace{-0.5em}
In this paper, we analyse the effects of domain, language, dataset size, multilinguality and phonetic information sharing in SSL pre-trained models for the downstream ASR fine-tuning setup. We study the impact of these factors both with and without continued pre-training of the SSL models. We think our work will be useful to the speech community, guiding the use of openly available pre-trained models and promoting efficient model re-use.
As part of future work, we would like to devise better language and domain adaptation methods for low-resource pre-training regimes.

\vfill\pagebreak



\bibliographystyle{IEEEbib}
\bibliography{strings}

\end{document}